\documentclass{article}

\usepackage[utf8]{inputenc}
\usepackage[english]{babel}
\usepackage{authblk}
\usepackage{breakcites}
\usepackage{booktabs,hyperref}
\usepackage{amssymb,amsmath,amsthm,twoopt,xargs,mathtools,cleveref}
\usepackage{times,dsfont,ifthen}
\usepackage{fancyhdr,xcolor}
\usepackage{xr,placeins}

\title{Joint self-supervised blind denoising and noise estimation}
\date{}

\author[$\star$]{Jean Ollion}
\author[$\amalg$]{Charles Ollion}
\author[$\dag$]{\'Elisabeth Gassiat}
\author[$\top$]{Luc Leh\'ericy}
\author[$\ddag$]{Sylvain Le Corff}

\affil[$\star$]{{\small SABILab, Die. \url{sabilab.fr}}}
\affil[$\amalg$]{{\small CMAP, \'Ecole Polytechnique, Institut Polytechnique de Paris, Palaiseau.}}
\affil[$\dag$]{{\small Universit\'e Paris-Saclay, CNRS, Laboratoire de math\'ematiques d'Orsay, 91405, Orsay.}}
\affil[$\top$]{{\small Laboratoire J. A. Dieudonn\'e, Universit\'e C\^ote d'Azur, CNRS, 06108, Nice.}}
\affil[$\ddag$]{{\small Samovar, T\'el\'ecom SudParis, d\'epartement CITI, TIPIC, Institut Polytechnique de Paris, Palaiseau.}}

\usepackage{geometry}
\newcommand{\interval}[2]{\mathopen{[}#1\,;#2\mathclose{]}}

\begin{document}

\maketitle

\begin{abstract}
We propose a novel self-supervised image blind denoising approach in which two neural networks jointly predict the clean signal and infer the noise distribution.
Assuming that the noisy observations are independent conditionally to the signal, the networks can be jointly trained without clean training data. Therefore, our approach is particularly relevant for biomedical image denoising where the noise is difficult to model precisely and clean training data are usually unavailable. Our method significantly outperforms current state-of-the-art self-supervised blind denoising algorithms, on six publicly available biomedical image datasets. We also show empirically with synthetic noisy data that our model captures the noise distribution efficiently. Finally, the described framework is simple, lightweight and computationally efficient, making it useful in practical cases.
\end{abstract}

\section{Introduction}
\label{sec:introduction}

Image denoising is a well-known Computer Vision task designed to restore pictures taken in poor conditions. In scientific imagery (microscopy, astronomy, etc.) for instance,  the optical setting may produce very noisy images, which limits their interpretability or their automatic processing.

Formally, image denoising is the process of recovering a clean signal $X$ given an observation $Y$ corrupted by an additive noise $\varepsilon$. Classical denoising approaches are model-driven in the sense that they rely on strong assumptions on the noise distribution or on the structure of the signal but are often limited by the relevance of these assumptions.
Recently, efficient data-driven methods have emerged. Most of them assume that pairs made of noisy data $Y$ associated with a clean signal $X$ are available in a supervised learning framework, see for instance \cite{weigert2017content}. In \cite{lehtinen2018noise2noise}, the authors have demonstrated that it is possible to train an efficient denoising method using only pairs of independent noisy measurements $(Y^1, Y^2)$ of the same signal. Such assumptions have also been used to solve deconvolution problems with repeated measurements as in \cite{delaigle2008deconvolution}. However, obtaining independent observations of the same signal is often unrealistic in practice.

Recent self-supervised methods have overcome this limitation \cite{batson2019noise2self,krull2018noise2void} by training a neural network to predict the value of a (corrupted) pixel $Y$ only using the noisy observations of the surrounding pixels. In such frameworks, the trained network extracts some local structure in the signal and therefore can be used as a denoiser. Such approaches are referred to as \textit{blind-denoising} as they only assume that the noises associated with different observations are independent and centered.
This is well suited to typical microscopy settings, in which the clean image is unavailable and the noise process is complex and not known.

These methods rely on training a function $f_\theta$ depending on an unknown parameter $\theta$, usually implemented as a convolutional neural network. Using only the noisy observations and a binary mask $M$, the objective is to minimize a self-supervision loss of the form $\theta\mapsto \sum_{i=1}^N \|f_\theta(Y^{masked}_i) - Y_i\|_2^2$,
where $Y^{masked}_i$ is an image in which the pixel $Y_i$ has been masked using $M$. This masking step is crucial to foster learning of local structure in the signal to predict the masked values.

While these methods are appealing in practice and result in efficient denoising functions, they suffer from several drawbacks:
\begin{itemize}
  \item It is not well understood why they are so effective in practice, i.e. what type of noise they are able to remove, and how sensible they are to the masking scheme for instance.
  \item They often suffer from high frequency denoising artifacts known as \textit{checkerboard pattern}.
\end{itemize}

To adress these issues, we introduce a novel self-supervised method, based on the joint training of two neural networks referred to as D-net (denoiser net) and N-net (noise net). Similar to previous works, the denoiser is a convolutional neural network and receives a masked input during training.
Our main contribution is to add the flexible N-net which recovers precisely the noise distribution during training, even for complex asymetric noises.
We derive this method from a novel mathematical modeling of the denoising problem, opening new avenues for better understanding of why self-supervised networks achieve remarkable results.

The contribution of this work can be summarized as follows.
\begin{itemize}
  \item We introduce a novel self-supervised blind-denoising method, modeling both the signal and the noise distributions.
  \item We show that the N-net recovers the noise distribution efficiently in varying experiments with synthetic and real noises.
  \item The proposed architectures outperform state-of-the-art algorithms over 6 standard microscopy datasets, without introducing denoising artifacts.
\end{itemize}

\section{Related work}
\label{sec:related}
\paragraph{Masking and J-invariance.}
The most typical class of denoising functions is chosen to be comprised of convolutional neural networks (CNNs), which are heavily parameterized functions and are not restricted to solve denoising problems. As an important consequence, a naive self-supervised loss without any masking would result in learning the identity function (i.e. the function outputing the noisy observation $Y$ if it is not masked in the input data), as the considered CNNs can typically implement it. Starting from this intuition, many of the related works can be viewed as different masking schemes. This has been described in the \textit{J-invariant} framework introduced by \cite{batson2019noise2self}: a \textit{J-invariant} function does not depend on a few selected dimensions $J$ of its input variables; typically this translates into a convolutional function which does not depend on the central pixel of the convolutional receptive field, but rather on the observations of neighboring pixels\footnote{Those functions excluding the central pixel are sometimes also called \textit{blind spot}, not to be confused with \textit{blind denoising} in which the noise process is not known.}.

The first masked self-supervised denoising methods were introduced by Noise2Void (N2V) \cite{krull2018noise2void}, in which $\{(Y^{masked}_i,Y_i)\}_{1\leqslant i\leqslant N}$ are sampled randomly in the picture, and masking consists in replacing $Y_i$ by a random observed value in its neighborhood, with a positive probability of replacing $Y_i$ by itself meaning that leaks in masking are introduced.

Noise2Self (N2S) \cite{batson2019noise2self} masking procedure differs from N2V in the sense that $\{(Y^{masked}_i,Y_i)\}_{1\leqslant i\leqslant N}$ are obtained with a fixed grid, and $Y_i$ is replaced by the average of the 4 direct neighboring observations.
In practice, the masking procedure has a strong impact on training: (i) improving masking schemes can improve denoising performance and (ii) as only masked pixels are used for training, typically representing a few percent of the image, this affects greatly the training efficiency.

The underlying CNN architecture implemented by these works is the U-net \cite{ronneberger2015u}, a typical convolutional autoencoder architecture, involving skip connections, which can reproduce fine grained details, while making use of higher-level spatially coarse information. While showing strong denoising performance in N2S and N2V, they however can produce \textit{checkerboard patterns}, which are high frequency artifacts that arise in the denoised results.
These works have been extended in DecoNoising \cite{goncharova2020} in which a Gaussian convolution is added after the neural network output to simulate microscope Point Spread Function.
This technique improves performances, however the deconvolved image (predicted image before the Gaussian convolution) displays even stronger checkerboard pattern.

Finally, \cite{broaddus2020removing} showed that when the noise has local correlations (for instance a  directional noise),  masking can be adapted to remove them - by masking adjacent pixels in the same direction as the noise spatial correlation for instance.

\paragraph{J-invariance without masking.}
Instead of masking specific pixels, it is possible to design specific convolutional operators to limit the receptive field, ensuring that the resulting function is \textit{J-invariant} by design. This was achieved in \cite{laine2019high} by introducing directional convolution kernels, each kernel has its receptive field restricted to a half-plane that does not contain the central pixel. The associated function then takes values which only depend on pixels in specific directions, ensuring that it does not depend on pixels in the opposite direction. One drawback is that the inference has to be performed four times, one in each direction.

More recently, \cite{lee2020noise2kernel} introduced a combination of specific convolution operators with dilation and strides, guaranteeing that the function is independent of the central pixel by design, therefore \textit{J-invariant}. It is interesting to note that with standard convolutions, a two layered network already cannot be made independent of the central pixel, which is why the authors had to rely on very specific convolutions.

The benefit of these architectures compared to the masking-based training is that all output pixels can contribute to the loss function as in conventional training, rather than just the masked pixels; and they do not require a carefully tuned masking procedure. However, they strongly constrain the network architecture, which can hinder the denoising performance or result in more expensive inference schemes.

\paragraph{Contribution of a noise model.}
The denoising literature includes few works which explicitly model the noise distribution, either by choosing \textit{a priori} a  family of distributions (e.g. a Gaussian noise), or by selecting a more flexible class of distributions.

The former is illustrated in \cite{laine2019high}, in which three types of corrupting noise are considered: Gaussian noise independent of the signal; Poisson-Gaussian noise, i.e. a Gaussian noise with variance scaling linearly with the signal value; finally impulse noise, i.e. a uniform noise. In each case, the noise parameters are either known or estimated with an auxiliary neural network. As the signal distribution and the noise distribution belong to a known parametric family, the noisy central pixel can be included at test time in order to improve performances. However, as the noise type has to be chosen \textit{a priori}, the method is restricted to known and synthetic noise types and therefore falls under the category of \textit{non-blind denoising} methods.

In \cite{krull2019probabilistic,prakash2020fully,2020DivNoising}, the authors make use of a more flexible noise model, which is a generic modelisation of the conditional distribution of the noise given the signal intensity and thus can better model real noises. In these works, noise models are approximated using 2D histograms of denoised and noisy observed values, either using additional calibration data (in that case the method is not fully self-supervised) or using a previously trained denoising function \cite{prakash2020fully}.
In the latter variant, the noise distribution is parametrized with a centered Gaussian mixture model with empirically designed constraints.
This increases the complexity of the method, as it requires several training procedures and calibration.
The denoising network is then trained to predict a whole distribution of $800$ possible noisy values instead of a single point estimate of each pixel, approximating the previously defined noise distribution at each pixel.

Finally, \cite{2020DivNoising} used the Variational Auto\-Encoder formalism, adding a pre-calibrated noise model in their architecture. This provides new interesting possibilities, as it can generate a diversity of denoised results, possibly interesting in creative processes. In the case of scientific images such as microscopy, the possible presence of visual artifacts or blurry results makes it less appropriate.

It is worth noting that supervised \textit{blind denoising} methods have used parametrized noise models, such as \cite{zhang2017beyond,yue2019variational}, which explicitly used large neural networks to model a complex noise, with even less assumptions (it can be slightly structured). Even though the algorithm proposed in \cite{yue2019variational} is able to train jointly a noise network and a denoiser, their modeling only works in a supervised setting, which does not apply in our setting.

\begin{figure}[ht]
\begin{center}
\centerline{\includegraphics[width=.8\columnwidth]{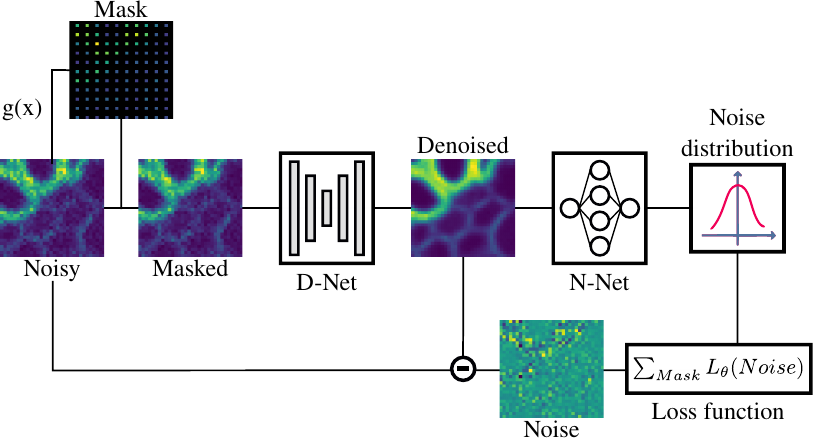}}
\caption{Training setup.
For each mini-batch, a random grid is drawn.
The masking function $x\mapsto g(x)$ is applied on each element of the grid, replacing the original pixels in the masked image.
The denoised image predicted by the D-net is fed to the N-net that predicts a noise distribution for each pixel.
The loss function is then computed on each element of the grid.
}
\label{fig:plumbing}
\end{center}
\vskip -0.2in
\end{figure}

\paragraph{Chosen approach.} The work of Laine et al. \cite{laine2019high} gave the intuition that a striclty J-invariant function lacks the information of the central pixel at test-time. On the contrary, methods such as N2V or N2S use the central pixel at test-time, but the dependency on the central pixel is not explicit and unknown. Instead of focusing on finding new strictly \textit{J-invariant} functions at test time, our approch rather emphasizes on designing an efficient masking procedure only at train-time. Our mathematical formulation enables the use of the central pixel at test time.
This also gives the flexibility to tune the masking to match structured noises, which we observed in 2 of the 6 considered datasets (see section~\ref{sec:masking}).
We also build upon the work of \cite{laine2019high, krull2019probabilistic} by designing a noise model which can be jointly trained alongside the denoiser (see Fig.~\ref{fig:plumbing}), and only requiring a single prediction per pixel: this results in a training and inference procedures that are simpler, more efficient and more stable.

\section{Model}
\label{sec:model}

Estimating a signal corrupted by additive noise  is a challenging statistical problem. In such frameworks, the received observation $Y$ is given by $Y = X + \varepsilon$,  where $X$ is the signal and $\varepsilon$ is the noise. A lot of works have been devoted to deconvolution where the aim is to recover the distribution of the signal based on the observations. It has been for instance applied in a large variety of disciplines and has stimulated a great research interest in signal processing \cite{moulines1997maximum,attias1998blind}, in image reconstruction \cite{kundur1996blind,campisi2017blind}, see also  \cite{meister:2009}. Recently, \cite{gassiat:lecorff:lehericy:2021} proved that it is possible to recover the signal distribution when $X$ has at least two dimensions and may be decomposed into two subsets of random variables which satisfy some weak dependency assumption. This identifiability result does not require any assumption on the noise distribution but illustrates that the components of the signal must be dependent to allow its identification. 

In this work, it is assumed that the observation $Y$ associated with $X$  is given by
\begin{equation}
\label{eq:def:Y}
Y = X + \sigma_{\theta_n}(X)\varepsilon\,,
\end{equation}
where $\varepsilon$ is a centered noise independent of $X$ and $\sigma^2_{\theta_n}$ is parameterized by a convolutional neural network, called N-net and with unknown weights $\theta_n$.
This contrasts with most common denoising algorithms where $\varepsilon$ is assumed to be a centered Gaussian random variable and  $x\mapsto \sigma^2_{\theta_n}(x)$ is either known and constant\footnote{From the perspective of our model, the approach of N2V and N2S is equivalent to considering that $x\mapsto \sigma^2_{\theta_n}(x) = 1$.} or has a Poisson-Gaussian shape i.e., scales with $\alpha x + \eta^2$.
As illustrated in Section~\ref{sec:results}, these assumptions do not usually hold, in particular when considering biomedical images, and they may have a severe impact on denoising performances.

In \cite{gassiat:lecorff:lehericy:2021}, $\sigma^2_{\theta_n}$ is assumed to be constant and the target signal is assumed to be weakly dependent to obtain identifiability of the noise and the signal  distributions. In \eqref{eq:def:Y}, we extend the model  proposed by \cite{gassiat:lecorff:lehericy:2021} by considering a state-dependent standard deviation and identifiability remains an open problem. However, we assume in this work that $X$ is dependent with the signal in the neighbooring pixels $\Omega_X$ so that heteroscedasticity is the only challenge to obtain identifiability of \eqref{eq:def:Y} which is left for future works.
In this work, we assume that $(X,\Omega_X)$ is a random vector with dependent variables and we propose to model the conditional mean of $X$ given $(Y,\Omega_Y)$ by a parametric function denoted by $\mu_{\theta_d}$ so that $\mathbb{E}[X|Y,\Omega_Y] = \mu_{\theta_d}(\Omega_Y,Y)$ where $\Omega_Y$ are the noisy observations of the signals in the neighborhood of $X$. The function $ \mu_{\theta_d}$ is parameterized by a convolutional neural network, called D-net and with unknown weights $\theta_d$.

A natural estimator of $X$ given the noisy observations is $\widehat X = \mu_{\theta_d}(\Omega_Y,Y)$.
During training this predictor $\widehat X$ cannot be used to estimate $\theta_d$ and $\theta_n$ as $\mu_{\theta_d}$ would learn to output the noisy observation $Y$ if it is not masked in the input data.
This is the reason why we adopt a making approach and assume during training that $\mu_{\theta_d}$ cannot use $Y$ as an input which must be replaced by an estimator.
In this framework, a genuine prediction of $Y$ is given by $\mathbb{E}[Y|\Omega_Y]$ which we estimate by $\mu_{\theta_d}(\Omega_Y,g(\Omega_Y))$ where $g$ is a known function. In the experiments below, we provide empirical evaluations that choosing $g(\Omega_Y)$ as the empirical mean of the noisy pixels in $\Omega_Y$ is a robust solution while other choices can be made straightforwardly.

Combining this with the additive model  \eqref{eq:def:Y} yields the following loss function associated with $N$ observations $(Y_1,\ldots,Y_N)$:
$$
\ell_{\theta}: (Y_1,\ldots,Y_N) \mapsto \frac{1}{N}\sum_{i=1}^N \ell_{\theta}(Y_i|\Omega_{Y_i})\,,
$$
where $\theta = (\theta_n,\theta_d)$ and
$$
\ell_{\theta}(Y_i|\Omega_{Y_i}) = \log(\sigma_{\theta_n}( \mu_{\theta_d}(\Omega_{Y_i},g(\Omega_{Y_i})))^2) + \left(\frac{Y_i-\mu_{\theta_d}(\Omega_{Y_i},g(\Omega_{Y_i}))}{\sigma_{\theta_n}(\mu_{\theta_d}(\Omega_{Y_i},g(\Omega_{Y_i}))}\right)^2\,. 
$$

An interesting  feature of our approach is that it can be extended straightforwardly to more complex noise distributions. As detailed above, model \eqref{eq:def:Y} is an extension of the model considered in \cite{gassiat:lecorff:lehericy:2021} where the authors establish that the noise distribution can be identified without any assumption. Full identifiability of model \eqref{eq:def:Y} remains an open question but we display in Section~\ref{sec:experiments} an example of non-Gaussian noises, and we propose an application with mixture models  to account for positive skewness which cannot be modeled with a single Gaussian distribution. In this context, each component of the mixture describing the distribution of $\varepsilon$ is a Gaussian distribution with signal-dependent standard deviation. The results provided in Section~\ref{sec:experiments} illustrate how such models improve denoising performance for asymmetrical noise distributions.

\section{Experiments}

\label{sec:experiments}
\subsection{Model Architecture}
\paragraph{D-net.}
The function $\mu_{\theta_d}$ is parametrized by a U-net. The main difference with the networks used in N2S and N2N is that we use upsampling layers with nearest-neighbor approximation instead of transpose convolutions, as we observed that transpose convolution tends to increase the checkerboard artefact.
Additional architecture and training information can be found in Section~\ref{si:implementation}.
The receptive field of this network is 35x35 pixels, which means that the network may use pixels from the neighorhood that are masked.
At test-time, we averaged the prediction of the image with the predictions of its transposed and flipped versions on each axis, which improves performances.

\paragraph{N-net.}
The function $\sigma_{\theta_n}: \mathbb{R} \to \mathbb{R}$ describing the local variance of the noise distribution is a fully-connected deep neural network with several hidden layers. This choice is motivated by the large expressivity of such a network, necessary to approximate complex noise distributions. In practice, it is applied to each pixel, so it is implemented efficiently as a fully convolutional network using only 1x1 convolutional layers. In the Gaussian Mixture Model (GMM) case, the network parametrizes a more complex distribution and therefore has several outputs: for a mixture of $N$ Gaussian distributions, there are $N$ variances, $N-1$ means (the last mean is computed to ensure that the resulting distribution is centered) and $N$ mixture weights parametrized by the N-net. The full architecture detail for both models are available in Section~\ref{si:implementation}.

\subsection{Datasets}
We train and evaluate our method on 6 publicly available datasets of microscopy images. In those datasets, ground truth ($X$) is estimated by averaging several observations ($Y$) of the same field-of-view (FOV).
This allows to have access to an estimation of the noise $Y-X$, which we refer to as \textit{real noise} in this article.

The 3 first datasets (\emph{PN2V-C}, \emph{PN2V-MN}, \emph{PN2V-MA}) have been published along with the PN2V method \cite{krull2019probabilistic}, each is composed of several observations of one single FOV.
For a fair comparison, we use the same training and evaluation sets as the authors: for each sample type the whole dataset is used for training, and only a subset of the FOV is used for evaluation (see Section~\ref{si:datasetxp} for details).

The 3 last datasets are the 3 channels of the W2S dataset \cite{zhou2020w2s} referred to as \emph{W2S-1}, \emph{W2S-2} and \emph{W2S-3}.
The dataset is composed of 120 FOV, the first 80 are used for training and the last 40 for evaluation (see Section~\ref{si:datasetxp} for more details).
Following the authors, for each FOV, only one observation is used for training and for evaluation, which better corresponds to a real setting where only one observation per FOV is available.

\subsection{Masking procedure}
\label{sec:masking}
Following \cite{batson2019noise2self}, we mask pixels along a grid and compute the loss only on masked pixels.
We obtained the best results by replacing the central value by the weighted average of the 8 direct neighbors with Gaussian weights ($\sigma=1$).
The drawback of masking along a grid is that pixels are masked at fixed relative positions with regards to the central pixel.
If grid spacing is too small, then too many masked pixels are present in the receptive field and perturb the performances, because the available information is reduced.
On the other hand, the larger the spacing, the less pixels are used for training, which reduces dramatically training efficiency.
In order to push the limits of this trade-off, we use a random dynamic spacing between 3 and 5 pixels, which allows to have relative positions of masked pixels that change randomly.
On average, $6.8\%$ of the image is masked.

Furthermore, we observed that datasets \emph{PN2V-C} and \emph{PN2V-MA} display axial correlation in the noise, for those datasets we adapted the masking procedure introduced in \cite{broaddus2020removing}: the replacement value was computed on a neighborhood excluding the neighbors along the correlation axis, and neighbors were masked along this axis, within an extent of 3 pixels\footnote{The masking extension can be determined easily in a self-supervised setup because the neural network tends to amplify the noise correlation, thus one can easily chose the smallest range for which the axial correlation disappears.}.

\subsection{Training}
Networks are trained using Adam optimizer with a learning rate of $4\cdot10^{-4}$, decreased by $1/2$ on plateau of 30 epochs until $10^{-6}$. We train networks for 400 epochs of 200 steps.
Training time is about 2 min per epoch on a NVIDIA Tesla P4.
We obtain better and more reproductible results using the weights of the trained model at the last epoch instead of the weights of the model with the best validation loss, possibly because the loss is a bad proxy for the denoising performances. For that reason, we do not use a validation step.
Batch size is set to 1, and each batch is split into 100 (overlapping) tiles of 96x96 pixels.
Tiles are augmented with random horizontal and/or vertical flip and/or a random rotation with an angle chosen within $(90^\circ, 180^\circ, 270^\circ)$\footnote{For the datasets with axial noise correlation, data-augmentation is only composed of combinations of flips to avoid axes transposition.}.

\subsection{Evaluation}
We compared denoised image to ground truth with the classical Peak Signal-to-Noise Ratio (PSNR) metric.
However PSNR is not highly indicative of perceived similarity, in particular it does not reflect similarity of high frequency information such as textures and local contrasts \cite{wang2004image}, that denoising methods tend to reduce. It is thus essential to have other metrics that them into account.
To address this shortcoming, we used Structural Similarity (SSIM) that take textures and edges into account \cite{wang2004image}, computed as in the original work.

\section{Results}
\label{sec:results}
\subsection{Noise estimation}

\begin{figure}[h!]
\begin{center}
\centerline{\includegraphics[width=.7\columnwidth]{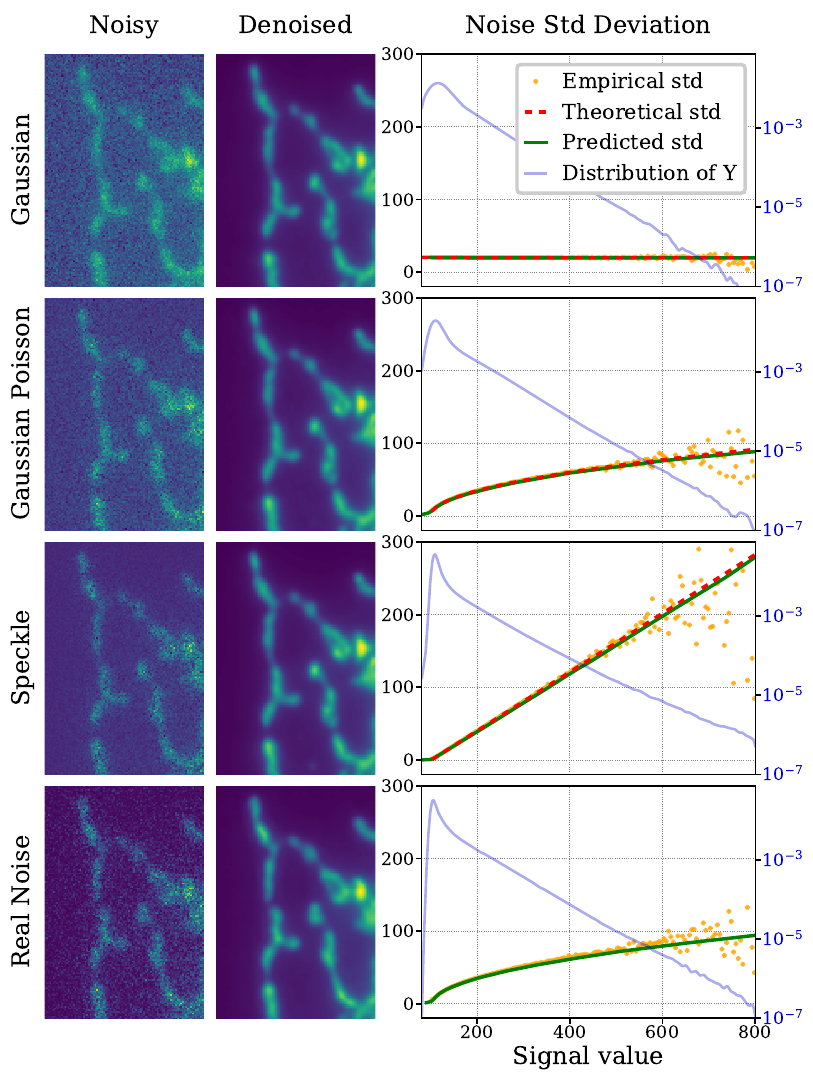}}
\caption[Noise estimation]{Noise estimation.
For 3 models of synthetic noise as well as the real noise, the plots display the empirical standard deviation of the noise $Y - X$, as well as the predicted standard deviation of the noise by the N-net\protect\footnotemark as a function of $X$.
Theoretical standard deviation of the noise is displayed for the 3 models of synthetic noise.
The empirical distribution of $Y$ is displayed in blue, in logarithmic scale.
Examples of noisy images and the corresponding predicted denoised images are displayed in columns \textit{Noisy} and \textit{Denoised}.}
\label{fig:noisestd}
\end{center}
\vskip -0.2in
\end{figure}
\footnotetext{Display range was shrinked in Y-axis for visualization purposes, excluding some points of the empirical standard deviation of the Speckle noise model.}
To evaluate the capacity of the N-net to capture blindly different noise distributions, we generated 3 datasets by adding synthetic noise to the ground truth of dataset \textit{W2S-1}, and we chose the parameters of the noise models so that PSNR of noisy images match the one of the original dataset (in a range of $\pm0.1$dB).
We used 3 classical noise models: additive Gaussian, Poisson-Gaussian (which is a good model for shot noise) and speckle (see Section~\ref{si:synthetic} for details).
Empirical and predicted distributions of the noise standard deviation are illustrated in Fig.~\ref{fig:noisestd}.
One of the most striking result of this experiment is that for the 3 cases of synthetic noise, the predicted standard deviation provided by the N-net is a very sharp approximation of the known theoretical standard deviation.
It shows in particular that our method is able to capture the different noise distributions even in areas where signal is rare.

\subsection{Improving estimation on real noise}
\label{sec:exp:real:noise}
\begin{figure}[h!]
\begin{center}
\centerline{\includegraphics[width=.7\columnwidth]{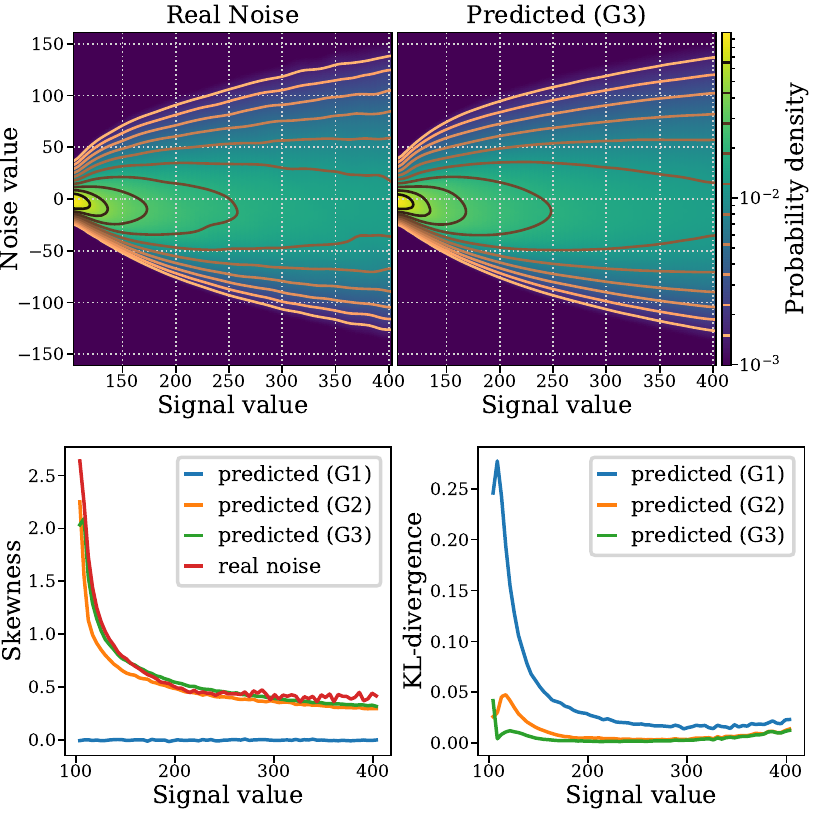}}
\caption{Real noise estimation for dataset \textit{W2S-1}. \textbf{Upper-left}: empirical distribution of the noise $Y - X$ as a function of $X$.
The probability density is normalized for each signal value bin.
\textbf{Upper-right}: corresponding predicted noise distribution for a 3-component-GMM.
\textbf{Lower-left}: Skewness of real and predicted noise distribution as a function of $X$, estimated with Pearson's moment coefficient of skewness\protect\footnotemark.
\textbf{Lower-rigth}:  Kullback–Leibler divergence between real noise distribution and predicted distribution generated by each model, as a function of $X$.
\textit{G1} stands for Gaussian model, \textit{G2} for a 2-component-GMM and \textit{G3} a 3-component-GMM.
}
\label{fig:skewness}
\end{center}
\vskip -0.2in
\end{figure}
\footnotetext{All the graphs are computed using regular signal value bins and excluding signal values greater to the $99.5\%$ percentile of the dataset so that there are enough observed samples in each bin to compute statistically significant metrics.}
We observed that contrary to the classical noise models considered in the denoising literature, real noise often displays a certain amount of skewness, as illustrated in Fig.~\ref{fig:skewness}.

\begin{table*}[ht]
\caption{Evaluation of our method on 6 datasets with PSNR/SSIM metrics.
SSIM estimates structural similarity (sharpness).
Metrics computed on noisy images are displayed in the \textit{Noisy} columns. For DecoNoising (\textit{DN}) and N2V, PSNR are taken from \cite{goncharova2020} and SSIM are computed on prediction made by networks trained using the source code provided by the authors\protect\footnotemark. \textit{Gaussian} corresponds to the optimal Gaussian baseline defined in section~\ref{sec:perf}.}
\label{table:results}
\vskip -0.15in
\begin{center}
\begin{small}
\begin{sc}
\resizebox{\textwidth}{!}{\begin{tabular}{l@{\hskip 7.5pt}c@{\hskip 7.5pt}c@{\hskip 7.5pt}c@{\hskip 7.5pt}c@{\hskip 7.5pt}c@{\hskip 7.5pt}c@{\hskip 7.5pt}c}
\toprule
Dataset & Noisy & Gaussian & N2V & DN & Ours (G1) & Ours (G2) & Ours (G3) \\
\midrule
PN2V-C & $28.98$ / $0.7713$ & $34.92$ / $0.9409$ & $35.85$ / $0.9404$ & $36.39$ / $0.9483$ & $38.33$ / $0.9754$ & $38.47$ / $0.9738$ & $38.28$ / $0.9756$ \\
PN2V-MN & $28.10$ / $0.6836$ & $35.53$ / $0.9392$ & $35.86$ / $0.9419$ & $36.34$ / $0.9489$ & $39.08$ / $0.9776$ & $39.22$ / $0.9779$ & $39.18$ / $0.9780$ \\
PN2V-MA & $23.71$ / $0.3731$ & $34.07$ / $0.8739$ & $33.35$ / $0.8384$ & $34.04$ / $0.8633$ & $34.79$ / $0.8905$ & $34.68$ / $0.8880$ & $34.70$ / $0.8877$ \\
W2S-1 & $21.85$ / $0.3490$ & $33.87$ / $0.9326$ & $34.30$ / $0.9026$ & $34.90$ / $0.9169$ & $35.33$ / $0.9619$ & $35.27$ / $0.9623$ & $35.27$ / $0.9624$ \\
W2S-2 & $19.33$ / $0.2256$ & $32.27$ / $0.8531$ & $31.80$ / $0.8311$ & $32.31$ / $0.8524$ & $33.46$ / $0.8867$ & $33.48$ / $0.8871$ & $33.47$ / $0.8871$ \\
W2S-3 & $20.39$ / $0.2232$ & $34.66$ / $0.9013$ & $34.65$ / $0.8637$ & $35.09$ / $0.9051$ & $36.57$ / $0.9263$ & $36.60$ / $0.9269$ & $36.59$ / $0.9269$ \\
\bottomrule
\end{tabular}}
\end{sc}
\end{small}
\end{center}
\vskip -0.2in
\end{table*}

In order to be able to capture this aspect, we predict a Gaussian mixture model (GMM) instead of a simple Gaussian model as described in Section~\ref{sec:model}. Fig.~\ref{fig:skewness} shows that noise skewness is well described by the predicted model, and the noise distribution is better described by a GMM than by a single Gaussian.
This applies for all datasets and the equivalent figures can be found in Section~\ref{si:skewness}.
In this example, it is interesting to note that the Kullback–Leibler divergence between the empirical noise distribution and the predicted distribution (as a function of the signal value) is improved by considering a GMM instead of a uninomodal distribution.
This supports the use of our flexible N-net to capture a large variety of noise distributions (with mutltimodality and/or skewness) which can be observed in experimental datasets.
This comment paves the way to several perspectives for our work such as the design of statistically consistent model selection procedures to choose automatically the number of mixing components.
Such approaches have been proposed in more simple cases using for instance penalized maximum likelihood based algorithms.
This remains an open problem in our framework and we leave this topic for future research.

\subsection{Denoising performances}
\label{sec:perf}
\footnotetext{\label{note:n2v}Using no positivity constraint, and removing the convolution for N2V.}
\begin{figure*}[h!]
\vskip 0.2in
\begin{center}
\includegraphics[width=\columnwidth
]{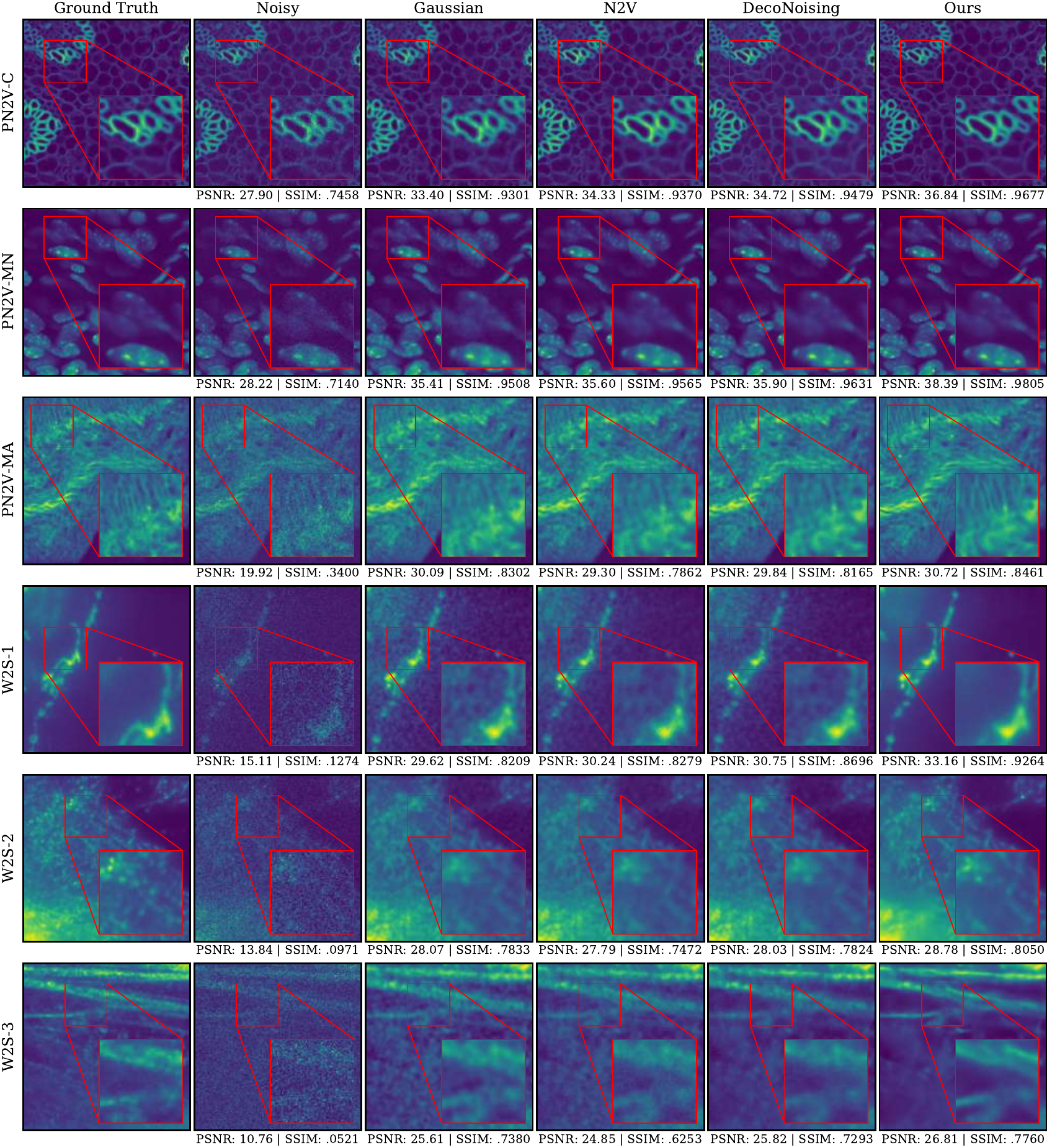}
\caption{Visual comparison of denoising on the considered datasets. For each dataset a $256$x$256$ portion of an evaluation image is displayed, on which metrics are computed and displayed below. For DecoNoising and N2V, images are predicted with networks trained using the source code provided with \cite{goncharova2020}\cref{note:n2v}. \textit{Gaussian} corresponds to the optimal gaussian baseline defined in section~\ref{sec:perf}.}
\label{fig:images}
\end{center}
\vskip -0.2in
\end{figure*}

We compared our method to 3 baselines: N2V, DecoNoising, which is the self-supervised blind denoising method that has shown best results on the datasets we considered, as well as one of the most simple denoising method: convolution by a Gaussian, whose standard deviation is chosen to maximizes the PSNR on the evaluation dataset.
We believe the latter makes a good reference, as it is one of the simplest denoising methods, and it removes noise efficiently but also other high-frequency information such as local contrasts.

The considered metrics are summarized in table~\ref{table:results}.
Our method significantly outperforms the 3 baselines both in terms of PSNR and SSIM on all datasets.
For the version predicting a simple Gaussian distribution, the average PSNR gain over DecoNoisng is $+1.42$dB.
This is also confirmed by the visual aspect, displayed in Fig.~\ref{fig:images}: our method produces images closer to the ground truth, smoother, sharper, more detailed and without visual artefacts.
Remarkably, our method performs significantly better than the supervised method CARE \cite{weigert2017content}, with an average PSNR gain of $+1.49$dB\footnote{Comparison with PSNR values reported in \cite{goncharova2020}}.

It is worth noting that considering mixture models improves the PSNR in two datasets out of six.
As mentionned in Section~\ref{sec:exp:real:noise} an optimal and data-driven choice of the number of components remains an open (and challenging) statistical problem but we believe that such experiments support future research in this direction.

\section{Discussion}
We introduced a novel self-supervised blind-denoising method modeling both the signal and the noise distributions. We believe its simplicity, performances and the interpretability of the noise distribution will be useful both in practical applications, and as a basis for future research.

First, future works could consider more complex families of noise distributions  such as structured or non-centered noises, that can also arise in real-life setups. In particular, \cite{lehtinen2018noise2noise} managed to remove very structured non-centered noises such as overlaid text. With stronger assumptions and architecture changes, it might be possible to capture such noises.

Second, more theoretical works could explore the model proposed in this work  (i) to obtain identifiability of model \eqref{eq:def:Y} and extend \cite{gassiat:lecorff:lehericy:2021} to state-dependent standard deviations and (ii) to establish rates of convergence for the proposed estimators.

Finally, it would also be interesting to understand the role of the central pixel at test time, as it has a significative impact on performance: it depends on the  masking procedure and the convolutional architecture, but the network is not trained explicitely to use it. Our mathematical modeling could be a good basis to study this specific dependency on the central pixel.

\section{Code}
Source code  will be available after peer review process.

\bibliographystyle{apalike}
\bibliography{blind_denoising}

\clearpage
\appendix
\section{Additional Implementation details}
\label{si:implementation}
\subsection{Networks and training}
\paragraph{D-Net architecture details.}
The architecture is based on U-net \cite{ronneberger2015u}.
We propose several changes from the original version: we do not crop the image and use zero-padding instead, we use 2 levels of contractions/expansions with 64 filters, expansions are performed by an upsampling layer with nearest-neighbor approximation directly followed by 2x2 convolution. We also add two layers of 1x1 convolution with 64 filters and ReLU activation at the end of the network, and set no activation function at the output layer.

\paragraph{N-Net architecture details.}
In the case of a Gaussian noise, the N-Net is composed 3 successive blocks, each block being composed of two 1x1 convolutions layers of 64 filters, each followed by a non-linear activation layer (alternatively tanh and leaky ReLU with alpha parameter set to $0.1$). A convolution 1x1 with a single channel followed by an exponential activation function is placed after the last block (to ensure that the predicted $\sigma$ is positive).

In the case where the N-net predicts a GMM with N components with weights $(\alpha_i)_{1\leqslant i\leqslant N}$, means $(\mu_i)_{1\leqslant i\leqslant N}$ and variances $(\sigma^2_i)_{1\leqslant i\leqslant N}$, the second block is connected to three distinct blocks, each connected to a convolution 1x1 with:
\begin{itemize}
  \item N channels, followed by an exponential activation function to predict $\sigma_{i}$.
  \item N channels, followed by a softmax activation to predict $\alpha_{i}.$\footnote{When $N=2$, only one channel is used and followed by a sigmoid activation function.}
  \item N-1 channels to predict the ditribution means $\mu_{i}$.
\end{itemize}
To ensure that the distribution is centered, the center of the last distribution is computed as
$$
\mu_{N} = - \frac{1}{\alpha_{N}} \sum_{i=1}^{N-1}{\alpha_{i} \mu{i}}\,.
$$

\section{Datasets}
\subsection{Experimental Datasets}
\label{si:datasetxp}
\paragraph{Datasets published along with the PN2V \cite{krull2019probabilistic}.}
\begin{itemize}
  \item \emph{Convallaria} dataset, referred to as \emph{PN2V-C} is composed of 100 images of size $1024$x$1024$. Evaluation subset is: $Y\in\interval{0}{512}, X\in\interval{0}{512}$.
  \item \emph{Mouse skull nuclei} referred to as \emph{PN2V-MN} is composed 200 images of size $512$x$512$. Evaluation subset is: $Y\in\interval{0}{512}, X\in\interval{0}{256}$.
  \item \emph{Mouse Actin} referred to as \emph{PN2V-MA} is composed of 100 images of size $1024$x$1024$. Evaluation subset is: $Y\in\interval{0}{1024}, X\in\interval{0}{512}$.
\end{itemize}

The \emph{PN2V-C} and \emph{PN2V-MA} datasets are acquired on a spinning disc confocal microscope and \emph{PN2V-MN} dataset is acquired with a point scanning confocal microscope.
Datasets can be downloaded from: \url{https://github.com/juglab/pn2v}

\paragraph{Datasets published in \cite{zhou2020w2s}.}

We used the 16-bit raw images kindly provided by the authors.
The dataset is composed of 120 FOV of 400 observations of size $512$x$512$ pixels.
The first 80 are used for training and the last 40 for evaluation.
Following the authors, for each FOV, only the observation of index 249 is used for training and evaluation
images are acquired with a electron-multiplying charge-coupled device camera on a wide-field microscope.
It can be downloaded here: \url{https://datasets.epfl.ch/w2s/W2S_raw.zip}

\paragraph{Normalization.}

For both datasets images were normalized using the modal value as center and the difference between modal value and $95\%$ percentile as scale factor, computed on the whole dataset.
This is relevant in fluorescence microscopy data where signal is often less abundant than background with proportion that vary among images and signal distribution often has a heavy tail towards high values.

\paragraph{Metrics.}
For the 6 chosen datasets, images are encoded in 16-bit.
PSNR is defined as 
$$
PSNR = 10 \log_{10}(d/\mathrm{MSE})\,,
$$
with $d$ the maximum possible pixel value of the image and $\mathrm{MSE}$ the mean squared error.
For 8-bit encoded images d is simply $255$, and for 16-bit images it would be $65635$ but this does not correspond to the actual possible range of microscopy data, thus the actual range of values of each ground truth image is used. This is also what is done in \cite{goncharova2020} as we obtain the same PSNR values for raw images.
The same applies for SSIM computation.

\subsection{Synthetic noise datasets}
\label{si:synthetic}
\begin{itemize}
  \item Additive gaussian: $Y = X + \varepsilon$ with $\varepsilon \sim \mathcal{N}(0, \sigma^2)$, $\sigma=20$.
  \item Poisson-Gaussian: $Y = X + (\alpha * (X-\underline{X}) + \eta^2 )^{1/2}\varepsilon$  with $\varepsilon \sim \mathcal{N}(0, 1)$, $\alpha=5, \eta=12$ and $\underline{X}$ being the minimal value of the ground truth on the whole dataset.
  \item Speckle: $X = X + (X-\underline{X})\varepsilon$  with $\varepsilon \sim \mathcal{N}(0, \sigma^2)$, $\sigma=0.405$ and $\underline{X}$ being the minimal value of the ground truth on the whole dataset.
\end{itemize}

\newpage
\section{Noise estimation}
\label{si:skewness}
\begin{figure}[ht]
\begin{center}
\centerline{\includegraphics[width=.7\columnwidth]{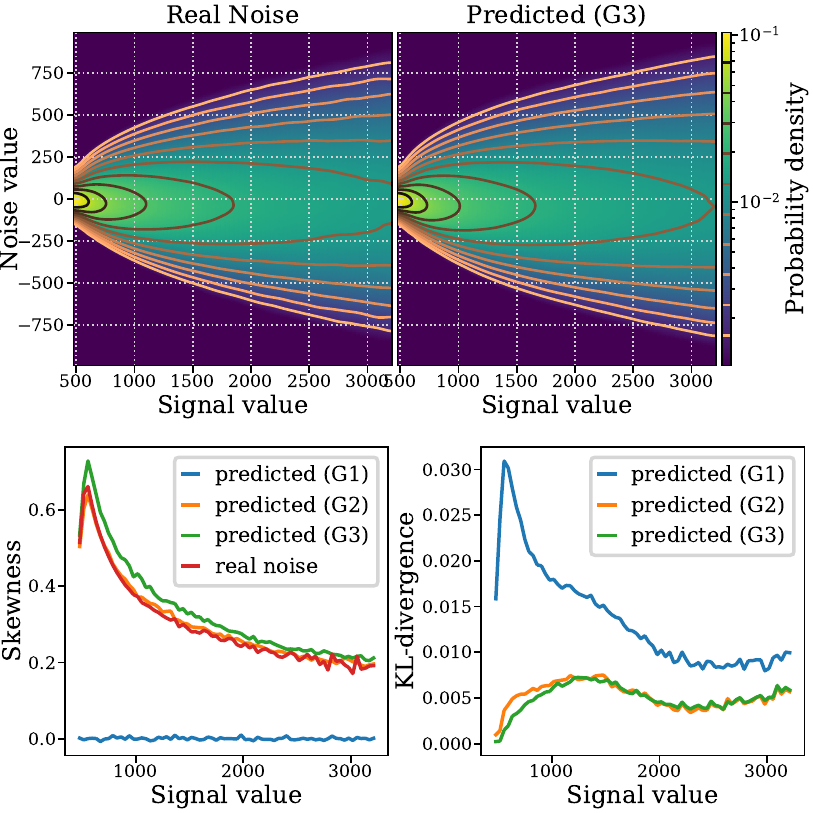}}
\caption{Real noise estimation for dataset \textit{PN2V-C}. See main text Fig~\ref{fig:skewness}.
}
\end{center}
\vskip -0.2in
\end{figure}

\begin{figure}[ht]
\begin{center}
\centerline{\includegraphics[width=.7\columnwidth]{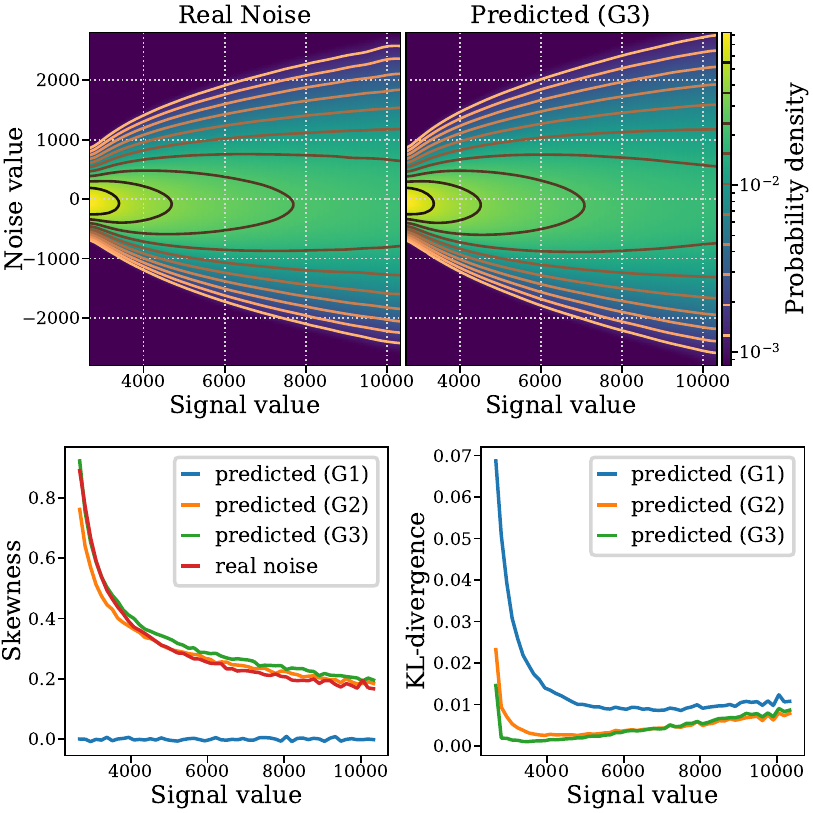}}
\caption{Real noise estimation for dataset \textit{PN2V-MN}. See main text Fig~\ref{fig:skewness}.
}
\end{center}
\vskip -0.2in
\end{figure}

\begin{figure}[ht]
\begin{center}
\centerline{\includegraphics[width=.7\columnwidth]{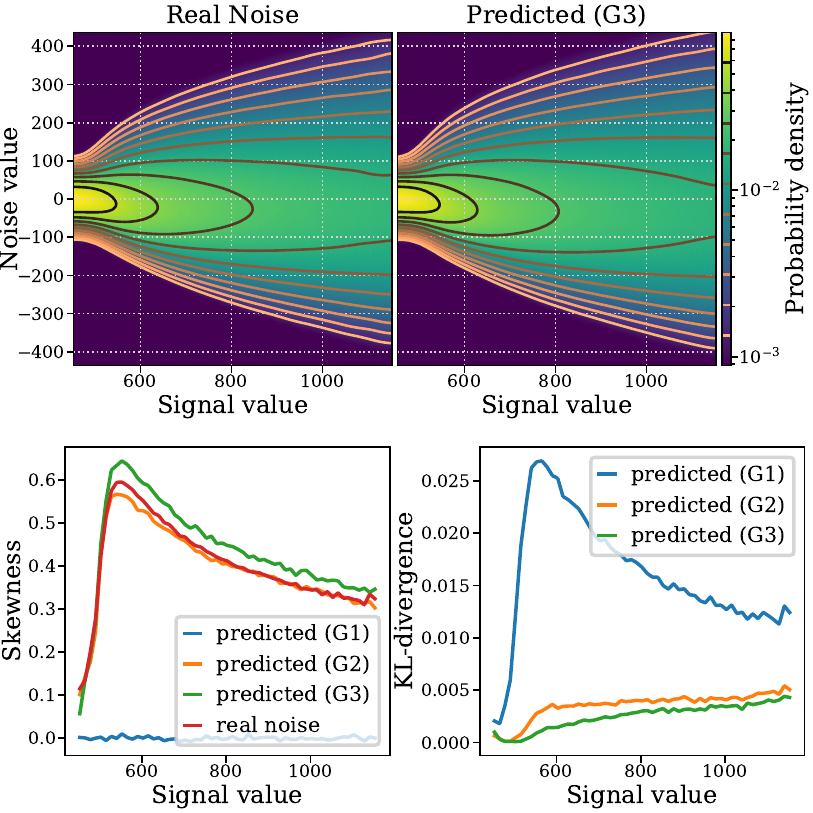}}
\caption{Real noise estimation for dataset \textit{PN2V-MA}. See main text Fig~\ref{fig:skewness}.
}
\end{center}
\vskip -0.2in
\end{figure}

\begin{figure}[ht]
\begin{center}
\centerline{\includegraphics[width=.7\columnwidth]{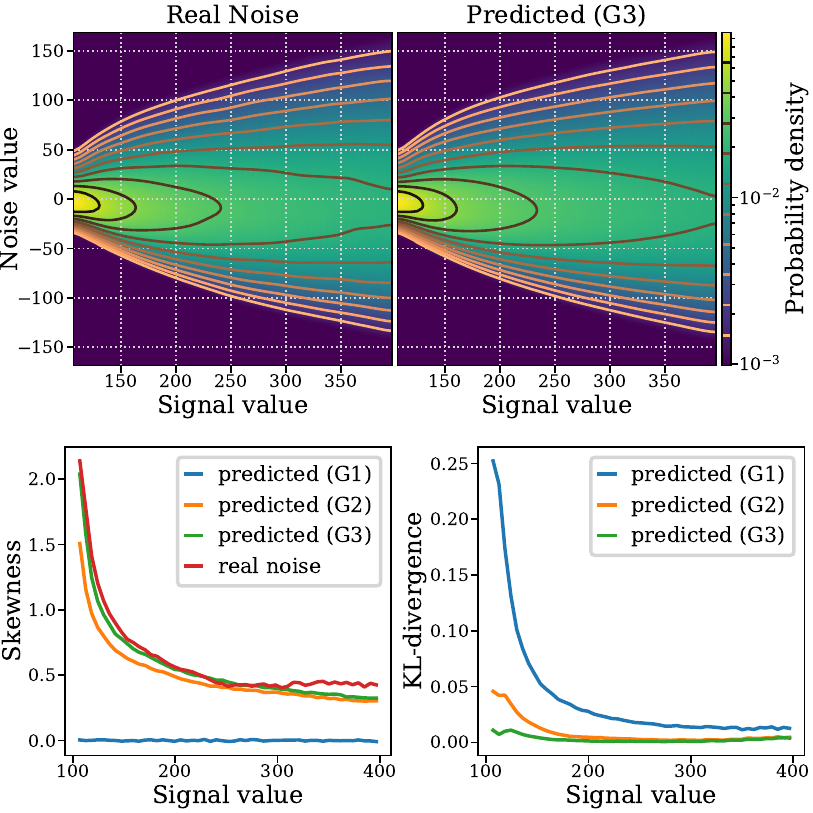}}
\caption{Real noise estimation for dataset \textit{W2S-2}. See main text Fig~\ref{fig:skewness}.
}
\end{center}
\vskip -0.2in
\end{figure}

\begin{figure}[ht]
\begin{center}
\centerline{\includegraphics[width=.7\columnwidth]{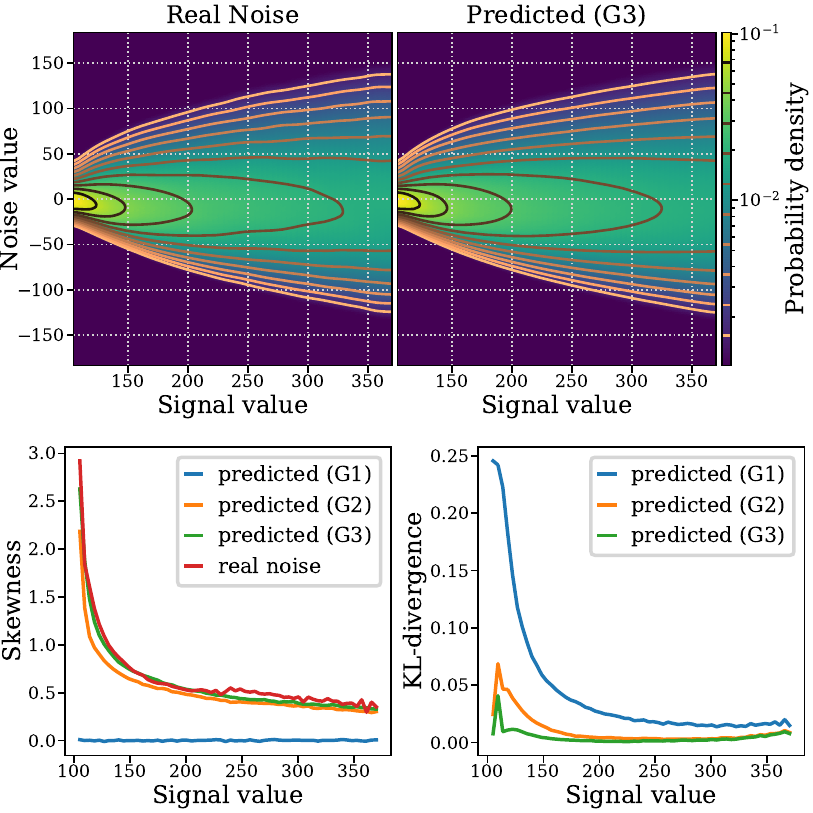}}
\caption{Real noise estimation for dataset \textit{W2S-3}. See main text Fig~\ref{fig:skewness}.
}
\end{center}
\vskip -0.2in
\end{figure}
\FloatBarrier

\end{document}